\documentclass[conference]{IEEEtran}
\IEEEoverridecommandlockouts
\usepackage{cite}
\usepackage{amsmath,amssymb,amsfonts}
\usepackage{graphicx}
\usepackage{colortbl}
\usepackage{gensymb}
\usepackage{multirow}

\usepackage{acronym}
\newacro{CNN}[CNN]{convolutional neural network}
\newacro{CCM}[CCM]{Color Conversion Matrix}
\newacro{RAE}[RAE]{recovery angular error}
\newacro{C3AE}[C3AE]{Color Constancy Convolutional AutoEncoder}
\newacro{GAN}[GAN]{Generative Adversarial Network}

\usepackage{algorithmic}
\usepackage{graphicx}
\usepackage{textcomp}
\usepackage{xcolor}
\def\BibTeX{{\rm B\kern-.05em{\sc i\kern-.025em b}\kern-.08em
    T\kern-.1667em\lower.7ex\hbox{E}\kern-.125emX}}

\begin{document}

\title{Color Constancy Convolutional Autoencoder\\
 \thanks{This  work  was  supported  by  the  NSF-Business Finland  Center  for
 Visual and Decision Informatics project (CVDI) sponsored by
 Intel Finland. }
 }



\author{ \IEEEauthorblockN{Firas Laakom$^{\dagger}$, Jenni Raitoharju$^{\dagger}$,  Alexandros Iosifidis$^{\star}$, Jarno Nikkanen$^{\dagger\dagger}$, Moncef Gabbouj$^{\dagger}$  }
\IEEEauthorblockA{$^{\dagger}$Tampere University, Faculty of Information Technology and Communication Sciences, Finland  \\
    $^{\star}$Aarhus University, Department of Engineering,  Denmark \\
     $^{\dagger\dagger}$ Intel  Corporation, Finland
    } 
}

\maketitle
\begin{abstract}
In this paper, we study the importance of pre-training for the generalization capability in the color constancy problem. We propose two novel approaches based on convolutional autoencoders: an unsupervised pre-training algorithm using a fine-tuned encoder and a semi-supervised pre-training algorithm using a novel composite-loss function. This enables us to solve the data scarcity problem and achieve competitive, to the state-of-the-art, results while requiring much fewer parameters on ColorChecker RECommended dataset. We further study the over-fitting phenomenon on the recently introduced version of INTEL-TUT Dataset for Camera Invariant Color Constancy Research, which has both field and non-field scenes acquired by three different camera models.
\end{abstract}

\section{Introduction}
\label{sec:intro}

Objects exhibit different colors under various light sources. The goal of color constancy algorithms is to remove this effect. This can be done by first estimating the color of the light source and using this illuminant estimate to transform the image as if it was taken under a neutral white light source. The aim of this transformation is not to scale the brightness level of the image, as color constancy methods only correct for the chromaticity of the light source.

Suppose we have the color of the unknown light source $ I( \lambda)$, the surface reflectance at location (\textit{x,y}),  $R(x,y,\lambda)$, and the camera sensitivity function  $S_i(\lambda)$, where $i$ is the color channel, i.e., $i \in (R,G,B)$.  Then the measured image color values $ \rho_i(x,y) $  at every pixel $(x,y)$ can be expressed as
\begin{equation}
            \rho_i(x,y) =  \int I(x,y, \lambda) R(x,y,\lambda) S_i(\lambda) d \lambda,  
\end{equation}
where $\lambda$ is the wave length. 
Color constancy methods aim to estimate the color $I(x,y)$ of the scene illuminant, i.e., the projection of $I( x,y,\lambda)$ on the sensor spectral sensitivities:
\begin{equation}
        I(x,y) =  \int I(x,y, \lambda) S_i(\lambda) d \lambda. 
\end{equation}
This problem is usually simplified by assuming a uniform light source color across the scene, i.e., $I(x, y) = I$. 

Deep neural networks have been recently extensively used  to approximate illumination and have often led to state of the art performance \cite{22,44,DSN,Das2018ColorCB,Barron2015ConvolutionalCC} across multiple datasets \cite{Hemrit2018RehabilitatingTC,47,17}. However, most of these supervised approaches were evaluated with a test data that is very similar to the training data, i.e., usually the training set and test set are acquired with the same camera models and they have similar types of scenes. In this paper, we highlight some limitations of the supervised approaches by taking the testing scenario to the extreme. We show that supervised models usually when  trained  on images  from  a  single  camera  and  a  single  scene type end up  learning  the  parameters  of  that  camera and scene and are not able to generalize effectively across other cameras and scenes. The over-fitting problem is a general issue in deep learning and it has been studied extensively. Erhan et al. \cite{erhan,erhan2} suggested that unsupervised pre-training makes it possible to obtain solutions that are similar in terms of training error but substantially better in terms of test error. They suggested that unsupervised pre-training has a dual effect both in helping optimization to start in better parameter space basins of attraction and as a kind of regularizer for the network.

State of the art color constancy \acp{CNN} based approaches \cite{44,DSN,Das2018ColorCB} usually use the first convolutional layers of a pre-trained model, e.g., SqueezeNet \cite{squueze}, AlexNet \cite{NIPS2012_4824}. While convolutional layers are proven to be effective in color constancy \cite{22,44,Barron2015ConvolutionalCC} in general, these pre-trained networks are originally trained for a classification task. Classification tasks benefit from being agnostic to illumination color. This makes their usage in color constancy counter-intuitive as illumination information must be preserved by the first layers to be able to detect it. Autoencoders provide a promising paradigm to use unsupervised pre-training for the color constancy context. A trained convolutional autoencoder can uncover the underlying structure of image chromaticities by learning over large numbers of unlabeled images that can be collected, for example, from the Internet. This can help generalize to unseen scenes and cameras without the need of very deep networks.

In this paper, we propose two novel approaches based on unsupervised pre-training using autoencoders. In the first, we learn a common representations of images and then fine-tune the model to estimate the illumination. In the second approach, we combine the two steps into one using a composite objective function which allows us to learn to reconstruct and, at the same time, regress to the illumination. 

\section{Related work}
\label{sec:related}

Typically, color constancy algorithms are divided into two main categories, namely unsupervised methods and supervised methods. The former involve methods with static parameters settings which are based on low-level statistics \cite{2,4,5,v4,v5} and methods using physics-based dichromatic reflection model \cite{3,8,11,v2}, while the latter involve data-driven approaches that learn to estimate the illuminant in a supervised manner using labeled data. 
Supervised methods can be further divided into two main categories: characterization-based methods and training-based methods. The former involve characterization of camera response in one way or another, such as Gamut Mapping \cite{64}, which assumes that in a real world scenario, for a given illuminant, only a limited number of colors can be observed. The latter involve methods that try to learn illumination directly from the scene \cite{v1,v3,45,34}. One group of training-based methods considers different illumination estimation approaches and learns a model that uses the best performing method or a combination of methods to estimate the illuminant of each input based on certain scene characteristics \cite{34}.  Another group of learning-based methods uses deep learning based approaches to solve the illumination estimation problem. 

The first attempt to use \acfp{CNN} for solving the illuminant estimation problem was done by  Bianco et al. \cite{22}, where they adopted a \ac{CNN} architecture operating on small local patches to overcome the limited number of training images available. In the testing phase, a map of local estimates is pooled to obtain one global illuminant estimate. For this approach, median pooling was shown to outperform other types of pooling techniques. Shi et al. \cite{DSN} proposed a network with two interacting sub-networks  to  estimate the illumination. One sub-network, called hypotheses network, is employed to generate multiple plausible illuminant estimations depending on the patches in the scene. The second sub-network, called the selection network, is trained to select the best estimate generated by the first sub-network. Das et al. formulated the illumination estimation task as an image-to-image translation task \cite{Das2018ColorCB} they used a \ac{GAN} to solve it. Barron \cite{Barron2015ConvolutionalCC}   reformulated the problem of color constancy as a 2D spatial localization task, in order to directly learn how to discriminate between correctly white-balanced images and poorly white-balanced images. Another CNN-based approach was proposed by Hu et al \cite{44}. They introduced a novel pooling layer, namely Confidence-weighted pooling layer in an end-to-end learning process. In their approach, patches in an image can carry different confidence weights according to the value they provide for color constancy estimation. In this deep model, pre-trained layers from SqueezeNet \cite{squueze} and AlexNet \cite{NIPS2012_4824} were used.

\section{Proposed approach } \label{Proposed}
While the current state of the art CNN-based methods use very deep models with convolution layers of a pre-trained model, we argue that unsupervised pre-training of a convolutional autoencoder may avoid overfitting without the need to go very deep. Training a Convolutional AutoEncoder (CAE) to reconstruct images and using it to estimate the illumination will allow us to use unlabeled data and thus obtain better parameters for the trained network. Learning to regenerate a large number of images from different cameras and sources will result in a model that will be more camera and scene invariant. We propose two approaches based on autoencoders named \acf{C3AE} \textit{fine-tuned} and \textit{\ac{C3AE} composite-loss}.

\textit{\ac{C3AE} fine-tuned } is a two-step approach. In the first step, an autoencoder is trained to reconstruct both labeled and unlabeled images to learn a latent representation for them using the binary cross-entropy loss. In the second step, the encoder part is fine-tuned to estimate the illumination using the  \acf{RAE} as the loss function. \ac{RAE} is a typical error measure in color constancy (see Section \ref{RAE}). 

In \textit{\ac{C3AE} composite-loss} approach, we combine the two steps of \textit{\ac{C3AE} fine-tuned} into one semi-supervised process. We train an autoencoder with a code size (middle layer) composed of only three neurons and  we reconstruct the images (labeled and unlabeled) while forcing at the same time the middle layer to regress to the desired illumination for the labeled samples. For this purpose, we modify the loss function of the autoencoder in the following manner:  
\begin{multline} \label{eq2}
\mathcal{L}_{new}(\mathcal{D\cup D'})=  \alpha * \frac{1}{ |\mathcal{D\cup D'}| } \sum\limits_{x \in \mathcal{D\cup D'}} \mathcal{L}(x, \tilde{x})\\
 +   (1 - \alpha) \frac{1}{|\mathcal{D}|} \sum\limits_{ x \in \mathcal{D}} RAE(\rho^{gt},\rho^{Est}) /90 ,
\end{multline}  
where  $\mathcal{D}$ is the labeled domain,  $\mathcal{D'}$ is the unlabeled domain, $|.|$ is the cardinality operator, i.e., number of elements in a set, $\mathcal{L}(x,\tilde{x})$ is the binary cross-entropy loss, $\ac{RAE}(\rho^{gt},\rho^{Est})$ is the angular loss (given in (4)) between the estimated illumination $\rho^{Est}$ in the bottleneck of the autoencoder and the ground truth illumination $\rho^{gt}$. The scaling by 1/90 makes the two losses of the same order of magnitude. The weight $\alpha$ is set as a hyperparameter. Intuitively, $\alpha$ encodes the weights of the two terms in the loss function. A small value means prioritizing the second term, i.e., learning to estimate the illumination, and a large value means prioritizing the first term, i.e, learning to reconstruct the images. To minimize Eq.~(\ref{eq2}), the autoencoder has to learn to reconstruct both at the labeled and unlabeled domain while matching the bottleneck as much as possible to the ground truth illumination for the labeled domain. In the last stage, the encoder part is fine-tuned using the labeled samples only.

\section{Experimental setup}
\subsection{Network architectures} 

We use a fully convolutional autoencoder which consists of four blocks of convolution, \textit{maxpooling}, and \textit{dropout} layers. The convolution filters are selected to be 32 of size 5*5 in the first two layers, 32 of size 4*4 in the third one, and 256 of size 3*3 in the fourth layer with an additional convolutional layer in the middle and the corresponding symmetric layers in the decoder.

For \textit{\ac{C3AE} fine-tuned}, the middle layer size is 50. The training is conducted with 1000 epochs and a batch size of 10.  For fine-tuning, in order to make the network suitable for illumination estimation,  we add two layers on top of the trained encoder: one of size 15 and the other of size 3. The fine-tuning is conducted with 1000 epochs and a batch size of 20. 
For \textit{\ac{C3AE} composite-loss}, the middle layer size is 3 and  $\alpha$ is equal to 0.5. 
Both trainings are conducted on image patches of size 64*64.
\subsection{Image datasets}

\subsubsection{ColorChecker RECommended dataset}\footnote{http://www.cs.sfu.ca/~colour/data/shi\_gehler/} 
ColorChecker RECommended dataset \cite{Hemrit2018RehabilitatingTC} is an updated version of Gehler-Shi dataset \cite{47} with a new proposed 'recommended' ground truth to use for evaluation. This dataset contains 568 high-quality mixed indoor and outdoor images acquired by two cameras: Canon 1D and Canon 5D. We use this dataset to evaluate the approaches in the first scenario, where  the test set is similar to the training set.

\subsubsection{INTEL-TUT2} 
INTEL-TUT2 is the second version of INTEL-TUT dataset \cite{17}. The main strength of this dataset is that it contains several camera models and several types of scenes organized separately. We use this dataset in the second training scenario, where the models are trained only with images acquired by one camera and containing one type of scene. The models are then tested on the other cameras and  scenes.

This publicly available\footnote{http://urn.fi/urn:nbn:fi:csc-kata20170901151004490662} dataset contains images taken with three cameras (namely Canon, Nikon, and Mobile). The images are divided into four sets: \textit{field} (144 images per camera), \textit{lab printouts} (300 images per camera), \textit{lab real scenes} (4 images per camera), and \textit{field2}. The last set \textit{field2} contains only images taken by Canon and it has in total 692 images. We use this last set for training and validation and the rest of the sets for the testing.

\subsubsection{Tiny ImageNet}
As unlabeled data, we used Tiny ImageNet\footnote{https://tiny-imagenet.herokuapp.com},  which is a smaller version of the original ImageNet \cite{imagenet_cvpr09}. We use 10k randomly selected images from this dataset. The diversity of ImageNet plays an essential role in this process. We believe that an autoencoder, which is trained to reconstruct this dataset, will encode a strong image dictionary. This will result in a stronger ability to generalize and help to build a robust illuminant estimator.  

\subsection{Evaluation procedure} \label{Evaluation}
For the first experiment, we used ColorChecker RECommended dataset. Similarly to \cite{22,44}, we used a three-fold cross validation on the folds provided with the dataset: for each run, one is used for training, one for validation, and the remaining one for testing

For the second experiment,  we used only Canon \textit{field2} set for training and validation (80\% for training and 20\% for validation). We constructed two test sets. The first one, referred to here as \textit{field}, contains all the field images taken by the other camera models, i,e., Nikon and Mobile. The second set, referred to here as \textit{non-field} contains all the non-field images acquired by Nikon and Mobile. This allowed us to test both scene and camera invariance of the models.

As in INTEL-TUT2 dataset different camera models are used, the variation of camera spectral sensitivity needs to be discounted. For this purpose, we utilize Color Conversion Matrix (CCM) based preprocessing\cite{25} to learn 3*3 CCM matrices for each camera pair.

For all the comparative experiments, data augmentation was performed as specified in the original works \cite{22,44}. For our models, we first downscaled the color constancy dataset images to 1920*1080 and randomly cropped 64*64 patches of these downscaled images. The crops were rotated by a random angle between -30\degree and +30\degree and, while training, we rescaled the patches and the corresponding ground truths by random RGB values in the range of [0.8, 1.2]. In testing, the images were first downscaled by 50\% in both axes and then 5 random 64*64 patches were selected from the image. This allowed us to generate a map of local estimates. We took the median of these estimates as the global illumination estimate.

 \subsection{Loss and evaluation metrics} \label{RAE}
For better insights into the robustness of the proposed methods, we report the mean of the top 25\%, the mean, the median, Tukey's trimean, and the mean of the worst 25\% of the \acf{RAE} \cite{21} between the ground truth illuminant and the estimated illuminant: 
\begin{equation}
     \text{\ac{RAE}}(\rho^{gt},\rho^{Est})= \cos^{-1} ({ \frac{ \rho^{gt} \rho^{Est}}{\| \rho^{gt} \| \|\rho^{Est} \| } }), 
\end{equation}
where $\rho^{gt}$ is the ground truth illumination for an image and $\rho^{Est}$ is the estimated illumination.

\section{Experimental results}
\subsection{Results on ColorChecker RECommended dataset} \label{RecResults}
We first evaluated accuracy of the approaches on ColorChecker RECommended dataset as shown in Table \ref{tab:table1}. We provide results for the static methods Grey-World, White-Patch, Shades-of-Grey, and General Grey-World. The parameter values $n$, $p$, $\rho$ are set as described in \cite{5}. In addition, we compare with Pixel-based Gamut, Bright Pixels, Spatial Correlations and six convolutional approaches: Deep Specialized Network for Illuminant Estimation (DS-Net) \cite{DSN}, Bianco \ac{CNN}  \cite{22},  Fast Fourier Color Constancy \cite{46440}, Convolutional Color Constancy\cite{Barron2015ConvolutionalCC}, Fully Convolutional Color Constancy With Confidence-Weighted Pooling (FC4) \cite{44}, and Color Constancy GANs (CC-GANs) \cite{Das2018ColorCB}.

\begin{table*}[h]

\renewcommand{\arraystretch}{1}

\caption{Results on ColorChecker RECommended dataset}
	\label{tab:table1}
\centering

\begin{tabular}{l|c|cccccc} 

\multirow{2}{*}{Method}                                    & \multicolumn{2}{c}{Type}                                        & \multicolumn{1}{l}{\multirow{2}{*}{Best 25\%}} & \multicolumn{1}{l}{\multirow{2}{*}{Mean}} & \multicolumn{1}{l}{\multirow{2}{*}{Med.}} & \multicolumn{1}{l}{\multirow{2}{*}{Tri.}} & \multicolumn{1}{l}{\multirow{2}{*}{Worst25\%}} \\
                                                           & \multicolumn{1}{c|}{statistic-based} & learning-based            & \multicolumn{1}{l}{}                           & \multicolumn{1}{l}{}                      & \multicolumn{1}{l}{}                      & \multicolumn{1}{l}{}                      & \multicolumn{1}{l}{}  
                                                           \\ \hline
                                                           
Grey-World \cite{4}  &\checkmark & --  &5.0 & 9.7 & 10 & 10 & 13.7 \\
White-Patch \cite{2}  &\checkmark&  -- & 2.2 & 9.1 & 6.7 & 7.8 & 18.9 \\

Shades-of-Gray \cite{shades}  &\checkmark &  --& 2.3 & 7.3 & 6.8 & 6.9 & 12.8 \\

General-gray world \cite{4} &\checkmark&  -- & 2.0 & 6.6 & 5.9 & 6.1 & 12.4 \\

Pixel-based Gamut \cite{644} & \checkmark& -- & 1.7 & 6.0 & 4.4 &4.9 &12.9 \\ 
Top-down \cite{high}  & \checkmark& -- & 2.3 & 6.0 & 4.6 & 5.0 & 10.2 \\ 
Spacial Correlations  \cite{chzcc2011} &\checkmark&  --& 1.9 & 5.7 & 4.8 & 5.1 & 10.9 \\ 

Bottom-up \cite{high} & \checkmark & -- & 2.3 & 5.6 & 4.9 & 5.1 & 10.2 \\ 

Edge-based Gamut \cite{644} & \checkmark& -- &  0.7 & 5.5 & 3.3 & 3.9 & 13.8 \\ 

		\hline

 CC-GANs (Pix2Pix) \cite{Das2018ColorCB} & --&\checkmark & 1.2 & 3.6 & 2.8& 3.1& 7.2 \\ 
 CC-GANs (CycleGAN) \cite{Das2018ColorCB} & --&\checkmark & 0.7 & 3.4 & 2.6& 2.8& 7.3 \\ 
 CC-GANs (StarGAN) \cite{Das2018ColorCB} & --&\checkmark & 1.7 & 5.7 & 4.9&5.2& 10.5 \\

FFCC (model Q) \cite{46440} & --&\checkmark &  0.3 &  2.0 &  1.1&  1.4& 5.1 \\ 
DS-Net \cite{DSN} & -- &\checkmark &  0.3 & 1.9 & 1.1 & 1.4 &4.8 \\
CCC\cite{Barron2015ConvolutionalCC} &  --&\checkmark  & 0.3 &  2.0 & 1.2 & 1.4 &4.8 \\
Bianco CNN \cite{22}& -- &\checkmark & 0.8 &2.6 &2.0 &2.1 & 4.0 \\
FC4(SqueezeNet)  \cite{44}& -- &  \checkmark  & 0.4 &  1.7 & 1.2 &  1.3 &   3.8\\

		\hline
\ac{C3AE}, fine-tuned  &  -- &\checkmark & 0.8 & 2.1 & 1.9 & 2.0 & 4.0\\
		\hline
\ac{C3AE},  composite-loss &  -- &\checkmark &  0.8 & 2.3 &  2.0 &  2.0 & 3.9\\

	\end{tabular}
\end{table*}

In this training scenario, training, validation, and test sets are similar in the sense that all of them contain images acquired with both camera models: Canon 1D and Canon 5D and various types of scenes. 
In this experiment, we note that learning-based methods usually outperform statistical-based methods across all error metrics. This can be explained by the fact that statistical approaches rely on some assumptions in their model. These assumptions can be violated in some testing samples and thus result in high error rates especially in terms of the worst 25\%.

In Table \ref{tab:table1}, we note also that  DS-Net, CCC, and  FFCC   achieve better error rates in terms of mean, median and trimean than our proposed  method \ac{C3AE} and its variants. But these methods are not stable and fail to generalize for many examples in the dataset. This can be seen through the worst 25\% error metric. The mean of the worse 25\% is bigger than 4.8\degree for these methods compared to 3.9\degree and 4\degree for our methods. Furthermore,  by comparing the number of parameters required by each model given in Table  \ref{tab:parameters}, we see that \ac{C3AE} achieves very competitive results, while using less than 1\% of the parameters  of DS-Net.

Table\ref{tab:table1} also shows that both of our proposed methods performs similar to Bianco \ac{CNN} w.r.t all metrics, except for the mean metric, where \acp{C3AE} outperform Bianco \ac{CNN}. The proposed approach shows competitive results compared to FC4, the error difference being less than $ 0.7 \degree$ for all the evaluation metrics, while using less than 10\% of the parameters. By comparing the number of parameters required by each model in Table  \ref{tab:parameters}, we see that \acp{C3AE} and its variant use less than 1\% of the parameters of FC4(SqueezeNet).

\textit{\ac{C3AE} fine-tuned} and \textit{\ac{C3AE} composite-loss} achieve similar results,  with \textit{\ac{C3AE} fine-tuned} performing better in terms of  the mean error metric and  \textit{\ac{C3AE} composite-loss} performing better in the mean of the worst 25\%.

\begin{table}[h]

\renewcommand{\arraystretch}{1}
 \centering	
	\caption{Number of parameters of different \ac{CNN}-based approaches}
	\label{tab:parameters}
	\begin{tabular}{l|r} 
		\hline
Method         &  \# parameters  \\	
\hline
Bianco  \cite{22} &   154k   \\
Fc4(SqueezeNet)   \cite{44} &   1.9M  \\
FC4 (AlexNet)   \cite{44}& 3.8M \\
DS-Net \cite{DSN} &   17.3M   \\
\hline
\ac{C3AE}, fine-tuned  &  146k \\
\ac{C3AE},  composite-loss &  146k \\

	\end{tabular}
\end{table}

\subsection{Results on INTEL-TUT2 dataset}

Table \ref{tab:table2} reports the comparative results and  the numbers of parameters for the \ac{CNN} based approaches: Bianco \ac{CNN}, FC4 (squuezeNet), \textit{\ac{C3AE} fine-tuned}, and \textit{\ac{C3AE} composite-loss} trained on INTEL-TUT2 dataset. To investigate the effect of pre-training on the performance of our approaches, we also provide results for \ac{C3AE} without pre-training. We provide the error metrics on three sets: the training set, the field, and non-field sets described in Section \ref{Evaluation}.

In this extreme scenario, the models are trained on \textit{field2} samples acquired with Canon. Then the testing is performed  on images acquired with other cameras and other type of scenes. For all the methods, we note a significant difference between the training errors and the test errors, i.e., most of the error metrics in both test sets have increased by a factor of 2-3 compared to the training errors. We note a slightly lower factor in our two proposed methods specially in terms of the worst 25\%. We also note that despite the fact that Bianco \ac{CNN} has a better training error rates than our methods, \textit{\ac{C3AE} fine-tuned} shows more generalization ability and outperforms Bianco \ac{CNN} in almost all test error metrics. \textit{\ac{C3AE} fine-tuned} shows competitive results compared to FC4 while using only 10\% of the parameters.

\begin{table}[h]
\footnotesize\setlength{\tabcolsep}{4pt}
\renewcommand{\arraystretch}{1}
 \centering	
	\caption{Results on INTEL-TUT2 dataset.}
	\label{tab:table2}
	\begin{tabular}{ll|ccccr} 
		\hline
Method  &  set         & Best \newline 25\% & Mean & Med. & Tri. & W. \newline 25\%  \\		\hline
\hline
 &   training   &  0.3 & 1.7 & 1.1 & 1.3 &  4.0  \\

Bianco  \cite{22}  &    field   & 1.1 &   4.5 &      3.7 &      3.8 &  9.2  \\
      &    non-field &    1.8 &    6.2 &   5.3 & 5.5 &     12.4  \\ 

      \hline
 &    training   &  0.6 & 1.6& 1.7 & 2.1  &  4.5  \\
FC4   \cite{44}    &   field   &       1.7 &      4.3 &       4.1 &         4.2 &       7.4  \\
(SqueezeNet)      &    non-field     &        1.5 &        4.8 &      4.2 &      4.3 &        9.0  \\ 	
\hline
      &   training   &   0.8 &  3.0 & 2.4 & 2.6 &  6.2  \\
\ac{C3AE} &    field   &        1.6 &        4.4 &         4.0 &    4.2 &          7.9  \\
fine-tuned       &   non-field     &        1.6 &        5.2 &       4.6 &       4.7 &        10.1 \\ 
   \hline 
 &    training   &   0.7 &  4.7 & 2.6 & 3.3 &  12.0  \\
\ac{C3AE} &    field   & 2.0 & 6.1 & 5.3 & 5.4 &    10.7\\
composite-loss   &    non-field     & 1.9 &    6.2 & 5.3 &    5.4 &  14.4  \\ 

\hline
&    training   &   0.5 &  1.6 & 1.6 & 1.9 & 10.6  \\
\ac{C3AE},  &    field   & 4.1 & 6.5 & 6.3 & 7.4 &  14.7\\
w.o pre-training  &   non-field     & 4.9 & 7.3 & 7.3 & 8.3 &  20.4  \\ 

	\end{tabular}
\end{table}

As we see in Table \ref{tab:table2}, unsupervised pre-training yields a much better generalization ability than semi-supervised pre-training in almost all error metrics. In comparison with the method without pre-training, we note that pre-training indeed helps and yields more robust methods. This can be explained by the fact that the autoencoder was trained with a diverse dataset containing images acquired with multiple cameras. This resulted in a robust initialization for the algorithms, which in turn resulted in models that can better generalize to different cameras and scenes.

\begin{figure}[h]
\centering
\includegraphics[width= 0.5\textwidth]{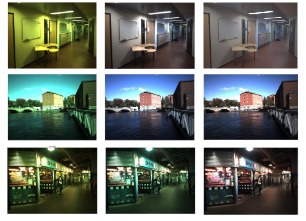}
\caption{Results of \ac{C3AE}, fine-tuned  for three samples from INTEL-TUT2 dataset. Each of the three column contains the input sample, the predicted result, and the ground truth, respectively.}
\label{viz_result}
\end{figure}

Figure \ref{viz_result} presents three samples from INTEL-TUT2 dataset, alongside their corresponding correction using \textit{\ac{C3AE}, fine-tuned} and their ground truth.

\section{Conclusion}
In this paper, illumination estimation algorithms were evaluated and compared on ColorChecker RECommended dataset. In addition, we tested the generalization ability of these algorithms in an extreme scenario with the second version of INTEL-TUT dataset, where color constancy approaches were trained using images only from one field set acquired with one camera and tested on images acquired with different camera models and on different scenes. We found that their performance drops significantly and they fail to some extent to generalize.

We proposed a method, \textit{\ac{C3AE}}, that exploits convolutional autoencoders and unsupervised pre-training to improve the generalization ability. With the proposed approach, we achieved comparable results to the state of the art methods using much fewer parameters.

Extensions of the proposed approach could include the use of other unsupervised pre-training techniques, such as variational convolutional autoencoders, in order to improve the generalization power from fewer examples.


\begin{thebibliography}{00}

\bibitem{22}
S.~Bianco, C.~Cusano, and R.~Schettini,
\newblock ``Color constancy using {CNN}s,''
\newblock in {\em IEEE Conference on Computer Vision and Pattern Recognition
  Workshops}, 2015, pp. 81--89.

\bibitem{44}
Y.~Hu, B.~Wang, and S.~Lin,
\newblock ``{FC}4: Fully convolutional color constancy with confidence-weighted
  pooling,''
\newblock in {\em IEEE Conference on Computer Vision and Pattern Recognition},
  2017, pp. 4085 -- 4094.
\bibitem{DSN}
W.~Shi, C.~C. Loy, and X.~Tang, ``Deep specialized network for illuminant
  estimation,'' in \emph{European Conference on Computer Vision}.\hskip 1em
  plus 0.5em minus 0.4em\relax Springer, 2016, pp. 371--387.

\bibitem{Barron2015ConvolutionalCC}
J.~T. Barron, ``Convolutional color constancy,'' \emph{2015 IEEE International
  Conference on Computer Vision (ICCV)}, pp. 379--387, 2015.


\bibitem{Das2018ColorCB}
P.~Das, A.~S. Baslamisli, Y.~Liu, S.~Karaoglu, and T.~Gevers, ``Color constancy
  by gans: An experimental survey,'' \emph{ Computing Research Repository}, 2018.



\bibitem{Hemrit2018RehabilitatingTC}
G.~Hemrit, G.~Finlayson, A.~Gijsenij, P.~Gehler, S.~Bianco, B.~Funt, M.~Drew,
  and L.~Shi,
\newblock ``Rehabilitating the colorchecker dataset for illuminant
  estimation,''
\newblock in {\em Color and Imaging Conference}, 2018, pp. 350--353.

\bibitem{47}
P.V. Gehler, C.~Rother, A.~Blake, T.~Minka, and T.~Sharp,
\newblock ``Bayesian color constancy revisited,''
\newblock in {\em IEEE Conference on Computer Vision and Pattern Recognition},
  2008, pp. 1--8.

\bibitem{17}
C.~{Aytekin}, J.~{Nikkanen}, and M.~{Gabbouj},
\newblock ``{A Data Set for Camera-Independent Color Constancy},''
\newblock {\em IEEE Transactions on Image Processing}, vol. 27, pp. 530--544,
  2018.







\bibitem{erhan}
D.~Erhan, P.A. Manzagol, Y.~Bengio, S.~Bengio, and P.~Vincent,
\newblock ``The difficulty of training deep architectures and the effect of
  unsupervised pre-training.,''
\newblock {\em Journal of Machine Learning Research - Proceedings Track}, vol.
  5, pp. 153--160, 2009.

\bibitem{erhan2}
D.~Erhan, Y.~Bengio, A.~Courville, P.-A. Manzagol, P.~Vincent, and S.~Bengio,
\newblock ``Why does unsupervised pre-training help deep learning?,''
\newblock {\em Journal of Machine Learning Research}, pp. 625--660, 2010.

\bibitem{squueze}
F.~Iandola, S~Han, M.~W.~Moskewicz, K.~Ashraf, W.~Dally, and K.~Keutzer,
\newblock ``Squeeze{N}et: {A}lex{N}et-level accuracy with 50x fewer parameters
  and $<$0.5{MB} model size,''
\newblock {\em CoRR - Computing Research Repository}, vol. abs/1602.07360,
  2017.

\bibitem{NIPS2012_4824}
A.~Krizhevsky, I.~Sutskever, and G.E. Hinton,
\newblock ``Image{N}et classification with deep convolutional neural
  networks,''
\newblock in {\em Advances in Neural Information Processing Systems 25}, pp.
  1097--1105. 2012.


\bibitem{2}
Al. Rizzi, C.~Gatta, and D.~Marini,
\newblock ``Color correction between gray world and white patch,''
\newblock in {\em Proceedings of SPIE - The International Society for Optical
  Engineering}, 2002, vol. 4662, p.~9.

\bibitem{4}
J.~Cepeda-Negrete and R.E. Sanchez-Yanez,
\newblock ``Gray-world assumption on perceptual color spaces,''
\newblock in {\em Image and Video Technology}, 2014, pp. 493--504.

\bibitem{5}
J.~van~de Weijer, T.~Gevers, and A.~Gijsenij,
\newblock ``Edge-based color constancy,''
\newblock {\em IEEE Transactions on Image Processing}, vol. 16, no. 9, pp.
  2207--2214, 2007.


\bibitem{v4}
A. Mecocci and G. Molinari, "Color recovery in outdoor environments: a novel integrated approach using retinex, gray world and stretching,"\emph{IEEE International Workshop on Multimedia Signal Processing},, 2004, pp. 75-78.

\bibitem{v5}
R. Lenz, Linh Viet Tran and P. Meer, "Moment based normalization of color images," \emph{IEEE International Workshop on Multimedia Signal Processing}, 1999, pp. 103-108.

\bibitem{3}
A.~Gijsenij, T.~Gevers, and J.~van~de Weijer,
\newblock ``Computational color constancy: Survey and experiments,''
\newblock {\em IEEE Transactions on Image Processing}, vol. 20, no. 9, pp.
  2475--2489, 2011.

\bibitem{8}
H.~C~Lee,
\newblock ``Method for computing the scene-illuminant chromaticity from
  specular highlights,''
\newblock {\em Journal of the Optical Society of America. A, Optics and image
  science}, vol. 3, pp. 1694--9, 1986.

\bibitem{11}
R.~Tan, K.~Nishino, and K.~Ikeuchi,
\newblock ``Color constancy through inverse-intensity chromaticity space,''
\newblock {\em Journal of the Optical Society of America. A, Optics, image
  science, and vision}, vol. 21, pp. 321--34, 2004.


\bibitem{v2}
F. Cogun and A. E. Cetin, "Object tracking under illumination variations using 2D-cepstrum characteristics of the target,"  \emph{IEEE International Workshop on Multimedia Signal Processing}, 2010, pp. 521-526.


\bibitem{64}
J.~Fr\"ohlich, A.~Schilling, and B.~Eberhardt,
\newblock ``Gamut mapping for digital cinema,''
\newblock in {\em SMPTE Annual Technical Conference Exhibition}, 2013, pp.
  1--11.

\bibitem{45}
V.C. Cardei, B.. Funt, and K.~Barnard,
\newblock ``Estimating the scene illumination chromaticity by using a neural
  network,''
\newblock {\em Journal of the Optical Society of America. A, Optics, image
  science, and vision}, vol. 19, pp. 2374--86, 2003.

\bibitem{34}
S.~Bianco, G.~Ciocca, C.~Cusano, and R.~Schettini,
\newblock ``Automatic color constancy algorithm selection and combination,''
\newblock {\em Pattern Recognition}, vol. 43, pp. 695--705, 2010.

\bibitem{v1}
Ling Hou, O. Au, Xiaopeng Fan and Jiantao Zhou, "Maximum-likelihood versus maximum a posteriori based local illumination and color correction algorithm for multi-view video," 2009 \emph{IEEE International Workshop on Multimedia Signal Processing}, 2009, pp. 1-4.

\bibitem{v3}
K. Bentolila and J. M. Francos, "Object pose estimation in the presence of local illumination changes using Scale Manipulation Transform," \emph{IEEE International Workshop on Multimedia Signal Processing}, 2009, pp. 1-6.

\bibitem{imagenet_cvpr09}
J.~Deng, W.~Dong, R.~Socher, L.-J. Li, K.~Li, and L.~Fei-Fei,
\newblock ``{ImageNet: A Large-Scale Hierarchical Image Database},''
\newblock in {\em IEEE Conference on Computer Vision and Pattern Recognition},
  2009.

\bibitem{25}
S.B. Gao, M.~Zhang, C.~Li, and Y.~Li,
\newblock ``Improving color constancy by discounting the variation of camera
  spectral sensitivity,''
\newblock {\em Journal of the Optical Society of America. A, Optics, image
  science, and vision}, vol. 34, no. 8, pp. 1448--1462, 2017.

\bibitem{21}
S.~D. Hordley and G.~D. Finlayson,
\newblock ``Re-evaluating colour constancy algorithms,''
\newblock in {\em International Conference on Pattern Recognition}, 2004, pp.
  76--79.

\bibitem{shades}
G.~Finlayson and E.~Trezzi,
\newblock ``Shades of gray and colour constancy,''
\newblock in {\em Color Imaging Conference}, 2004, pp. 37--41.

\bibitem{644}
G.~{Finlayson} and S.~{Hordley}, ``Improving gamut mapping color constancy,''
  \emph{IEEE Transactions on Image Processing}, vol.~9, no.~10, pp. 1774--1783,
  Oct 2000.




\bibitem{high}
J.~{van de Weijer}, C.~{Schmid}, and J.~{Verbeek}, ``Using high-level visual
  information for color constancy,'' in \emph{2007 IEEE 11th International
  Conference on Computer Vision}, Oct 2007, pp. 1--8.



\bibitem{46440}
J.~T. Barron and Y.-T. Tsai, ``Fast fourier color constancy,'' in \emph{CVPR},
  2017.




\bibitem{chzcc2011}
A.~Chakrabarti, K.~Hirakawa, and T.~Zickler, ``Color constancy with
  spatio-spectral statistics,'' \emph{IEEE Transactions on Pattern Analysis and
  Machine Intelligence}, 2012.


\bibitem{7298702}
D.~{Cheng}, B.~{Price}, S.~{Cohen}, and M.~S. {Brown}, ``Effective
  learning-based illuminant estimation using simple features,'' in \emph{2015
  IEEE Conference on Computer Vision and Pattern Recognition (CVPR)}, June
  2015, pp. 1000--1008.





\end{thebibliography}
\end{document}